\documentclass[journal]{IEEEtran}
\usepackage{amsmath,amsfonts}
\usepackage{algorithmic}
\usepackage{authblk}
\usepackage{hyperref}
\usepackage{xcolor}
\usepackage{doi}
\usepackage{array}
\usepackage[caption=false,font=normalsize,labelfont=sf,textfont=sf]{subfig}
\usepackage{textcomp}
\usepackage{stfloats}
\usepackage{url}
\usepackage{verbatim}
\usepackage{booktabs}
\usepackage{multirow}
\usepackage{graphicx}
\graphicspath{{Figures/PDF/}{Figures/PNG/}}
\usepackage[numbers,compress]{natbib}
\usepackage{orcidlink}
\def\BibTeX{{\rm B\kern-.05em{\sc i\kern-.025em b}\kern-.08em
    T\kern-.1667em\lower.7ex\hbox{E}\kern-.125emX}}
\usepackage{balance}

\begin{document}
\title{SAR2Struct: Extracting 3D Semantic Structural Representation of Aircraft Targets from Single-View SAR Image}

\author{Ziyu Yue,~\IEEEmembership{Graduate Student Member,~IEEE,} Ruixi You,~\IEEEmembership{Graduate Student Member,~IEEE,} and Feng Xu\raisebox{0.5ex}{\orcidlink{0000-0002-7015-1467}},~\IEEEmembership{Senior Member,~IEEE}
\thanks{This work was supported in part by ... \textit{(Corresponding author: Feng Xu)}}
\thanks{The authors are with the Key Laboratory of Information Science of Electromagnetic Waves (Ministry of Education), Fudan University, Shanghai 200433, China (e-mail: fengxu@fudan.edu.cn).}
}

\markboth{Journal of \LaTeX\ Class Files}%
{SAR2Struct: Extracting 3D Semantic Structural Representation of Aircraft Targets from Single-View SAR Image}

\maketitle

\begin{abstract}
To translate synthetic aperture radar (SAR) image into interpretable forms for human understanding is the ultimate goal of SAR advanced information retrieval. Existing methods mainly focus on 3D surface reconstruction or local geometric feature extraction of targets, neglecting the role of structural modeling in capturing semantic information. This paper proposes a novel task: SAR target structure recovery, which aims to infer the components of a target and the structural relationships between its components, specifically symmetry and adjacency, from a single-view SAR image. Through learning the structural consistency and geometric diversity across the same type of targets as observed in different SAR images, it aims to derive the semantic representation of target directly from its 2D SAR image. To solve this challenging task, a two-step algorithmic framework based on structural descriptors is developed. Specifically, in the training phase, it first detects 2D keypoints from real SAR images, and then learns the mapping from these keypoints to 3D hierarchical structures using simulated data. During the testing phase, these two steps are integrated to infer the 3D structure from real SAR images. Experimental results validated the effectiveness of each step and demonstrated, for the first time, that 3D semantic structural representation of aircraft targets can be directly derived from a single-view SAR image.
\end{abstract}



\section{Introduction}
\IEEEPARstart{S}{ynthetic} aperture radar (SAR) can acquire high-resolution microwave images of targets and environments from distance regardless of weather and daylight, making it an essential tool for Earth remote sensing~\cite{xu2016preliminary},~\cite{jia2024fast}. SAR imagery is known to be difficult to understand due to its unique imaging mechanism and the complex electromagnetic scattering phenomenon. To translate high-resolution multidimensional SAR data into interpretable forms for human analysis is the ultimate goal of SAR advanced information retrieval (AIR)~\cite{xu2016preliminary}, where 3D target reconstruction is one of the main pathways. However, existing methods mainly focus on 3D surface reconstruction or local geometric feature extraction of targets, neglecting the role of structural modeling in capturing semantic information. In recent years, impressive progress has been achieved in the structured representation of 3D shapes using deep learning. Yet no prior work has leveraged or extracted the 3D structural information of targets embedded in SAR images. In fact, high-resolution 2D SAR images provide rich spatial cues such as target structure and texture, which are crucial for 3D modeling. It poses new challenges in the extension of conventional optical 3D reconstruction methods to SAR-based 3D target reconstruction. For example, it is difficult to obtain a large and balanced dataset of SAR images representing various target categories, and the scarcity of corresponding ground truth data for the 3D shapes of these targets makes supervised training with deep learning models challenging. Moreover, the complexity and variability of image features exacerbate the problem. Targets in SAR images are often sparsely represented with point-like characteristics and are subject to speckle noise interference, making it complicated to learn effective features that are both discriminative and robust with limited data.

\begin{figure}[t]
	\centering
	\includegraphics[width=0.90\linewidth]{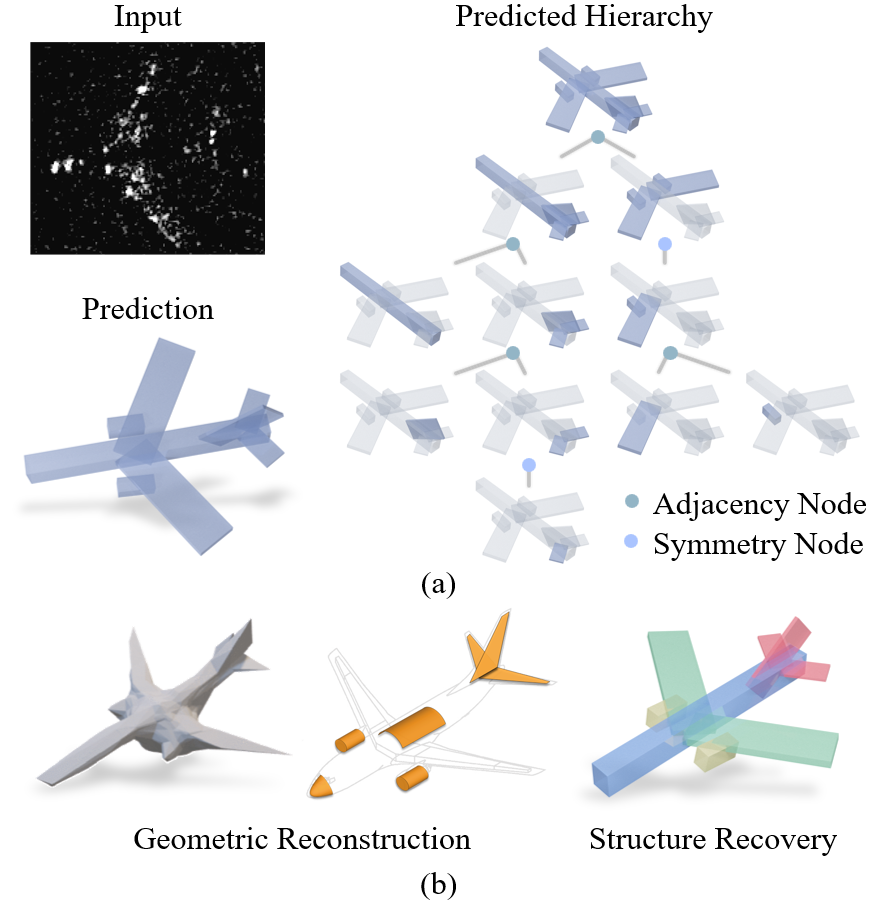}
	\caption{SAR aircraft structure recovery. (a) introduces SAR2Struct, which predicts the hierarchical 3D structure from an aircraft's SAR image. (b) highlights the difference between existing studies (left and middle) that focus on geometric features and ours (right) that adopts a structural perspective.}\label{fig:teaser1}
\end{figure}

The introduction of differentiable renderers~\cite{liu2019soft} has expanded the approach to 3D target reconstruction. Recent studies~\cite{fu2022differentiable},~\cite{fu2023extension} have probabilistically modeled the mapping and projection algorithm~\cite{xu2006imaging} used in SAR image simulation as a forward rendering model, minimizing the discrepancy between the target SAR image contours and rendered images. The 3D geometric surface of the target is then reconstructed through error backpropagation (BP) algorithms. These methods typically employ mesh representations of targets and focus primarily on forward and inverse scattering modeling between SAR images and targets' overall surface, overlooking the role of local geometric priors(Fig.~\ref{fig:teaser1}(b) – left).

In contrast, scattering center–based approaches for SAR target reconstruction place a stronger emphasis on geometric priors(Fig.~\ref{fig:teaser1}(b) – middle). These methods generally involve three steps: (1) constructing a parameterized scattering dictionary~\cite{potter1997attributed}; (2) reformulating the reconstruction task into a combination of scattering mechanism classification and dictionary parameter regression; and (3) assembling the reconstructed scattering components~\cite{li2016complex}. Although classical scattering mechanisms—such as planes, spheres, and trihedral corners—are employed as dictionary elements to provide local semantic insights, these methods overlook global structural priors, namely the relationships among different components.

Both of the aforementioned SAR 3D target reconstruction methods rely on forward scattering models—in essence, physically derived descriptors capturing either global or local scattering properties of the SAR image. While this physical perspective is crucial, the equally important structural perspective (Fig.~\ref{fig:teaser1}(b) - right) has been largely overlooked by existing methods. Therefore, the aim of this paper is to propose the SAR structure recovery task, illustrated in Fig.~\ref{fig:teaser1}(a), enriching the notion of SAR 3D reconstruction from a novel structural point of view.

This approach is intuitively appealing: human perception, cognition, and understanding of the world are inherently structured. Similar objects usually share similar structures, and their structurally analogous parts tend to fulfill similar functions and low-dimensional observational features. For example, the 3D shape of a chair typically comprises a base, seat, armrests, and backrest, whereas most fixed-wing aircraft incorporate a fuselage, wings, tailfins, and engines. The fuselage generates curved and multiple scattering, wings frequently exhibit edge diffraction, tailfins produce multiple scattering and edge diffraction, and engines often produce cavity scattering. From a structural perspective, SAR images can thus be modeled as observations of similar targets characterized by structural consistency and geometric diversity.

\begin{figure}[t]
	\centering
	\includegraphics[width=0.78\linewidth]{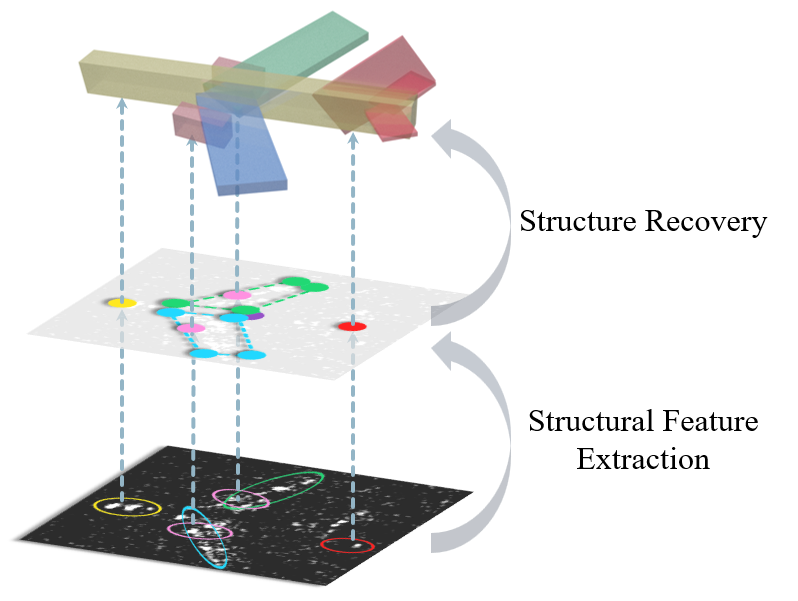}
	\caption{Implementation ideas of SAR structural recovery. Starting from SAR imagery (bottom), key semantic features are extracted (middle) and mapped to a 3D structure (top).}\label{fig:teaser2}
\end{figure}

Additionally, this approach offers practical advantages. From a physical perspective, an algorithm's performance depends on how accurately the physical model reflects real electromagnetic scattering and SAR imaging mechanisms. However, meeting this condition is challenging because SAR image features are complex and variable, inevitably causing discrepancies between physically simulated SAR images and their real counterparts. Consequently, features extracted from simulated images may fail to transfer to real-world data. In contrast, the commonality between simulated and real SAR data from a structural perspective is evident. By selectively extracting structural features from SAR images, rather than relying on spatially continuous features like geometry and texture, one can mitigate performance degradation when adapting to new datasets. Specifically, as shown in Fig.~\ref{fig:teaser2}, by introducing a structure descriptor with physical meaning as an intermediary, two sets of mapping relationships can be modeled separately, allowing for the training of 3D structure recovery using simulated data and mitigating the domain gap between simulated and real data.

3D shape representation is a fundamental problem in computer graphics. Researchers have pointed out that man-made objects often exhibit notable symmetry and hierarchy. By capitalizing on these two characteristics, a tree-based structural representation is widely used, known as Symmetry Hierarchy (SYMH)~\cite{wang2011symmetry}, which describes how different parts of a 3D shape are grouped via symmetry or assembled via adjacency, defining the computational logic of tree nodes. The development of deep learning has further promoted research into SYMH, e.g. the recursive neural networks (RvNN) designed to generate SYMH representations~\cite{li2017grass}.

This structured representation approach is especially compelling to us. Hierarchy provides high-level abstractions that highlight common features at a coarse granularity, while preserving fine-grained diversity for flexible representation. Meanwhile, symmetry provides compactness by removing redundant information—an essential step toward more accurate structural expressions. In this work, we adopt the SYMH representation and the recursive network structure to motivate the 3D structure recovery from aircraft SAR images. 

The main contributions of this paper are summarized as follows:
\begin{enumerate}
    \item We propose a new task of target 3D semantic structure recovery, which is to infer the components of a target and the structural relationships among them, specifically symmetry and adjacency, from a single-view SAR image. Compared to existing 3D reconstruction tasks, it pushes one step further towards the ultimate goal of SAR AIR, i.e. translation of SAR image into directly interpretable form for human understanding.
    \item To solve this new task, a novel two-step algorithmic framework based on structural descriptor transitions is developed, which decouples 2D component keypoint detection from 3D structure recovery, effectively addressing the scarcity of SAR data and the domain gap between simulated and real SAR images.
    \item Finally, an end-to-end semantic structural extraction tool is implemented which, for the first time, demonstrated on real SAR images that semantic 3D hierarchical structure representation of aircraft targets can be directly derived from a single-view SAR image.
\end{enumerate}


\section{Related Work}
\subsection{SAR 3D Reconstruction}
\subsubsection{Target surface reconstruction}
Reconstructing 3D shapes from single-view or multi-view images is a fundamental problem in computer vision. In the context of SAR target reconstruction, point clouds are frequently adopted as a 3D representation. Peng et al.\cite{peng2019generating} transformed MSTAR dataset~\cite{mstar2014} into optical images and obtained corresponding point cloud using a pre-trained model. Other works\cite{feng20213d},\cite{wang2023multi} have reconstructed 3D shapes of civilian vehicles. Additional efforts have focused on buildings\cite{han2023geometric} and airports~\cite{zhang2022probabilistic}. While point clouds fail to fully capture the surface geometry, meshes offer more powerful expressive capabilities. Consequently, Yu et al.\cite{yu2023lightweight} proposed a lightweight Pixel2Mesh model for reconstructing 3D targets. Other works\cite{fu2022differentiable},\cite{fu2023extension} integrated differentiable renderers into 3D mesh reconstruction pipelines. Furthermore, Qin et al.\cite{qin2024sar} learned the voxel representation of aircraft and vehicles. However, all these studies have not taken into account component descriptions of the 3D models and relationships between the components.

\subsubsection{Local geometric extraction}
Similar to methods such as~\cite{li2019supervised} that use primitive dictionary sets to represent 3D shapes, the parameterized scattering center model builds scattering dictionaries for classical geometry to describe radar objects, transforming 3D reconstruction into a parameter estimation problem. For example, the region decoupled algorithm~\cite{koets1999image} adopts a strategy of image segmentation and region-by-region estimation. Similarly, works such as~\cite{ding2018target},\cite{jing2020attributed} also employ approximate maximum likelihood (AML) estimation. Works such as~\cite{yang2020efficient},\cite{xie2022attributed} introduce sparse priors, using Newton's orthogonal matching pursuit (OMP) to tackle the sparse characteristics of target’s scattering field. Furthermore, other studies~\cite{chen2024reinforcement},~\cite{yue2024extraction} developed deep learning algorithms to extract image features and estimate parameters. Nonetheless, parameterized model approaches suffer from low accuracy in scattering mechanism classification, and currently struggle to produce representations that closely match the target’s 3D shape.

Compared with the methods discussed above, our work focuses on recovering the 3D structure from aircraft SAR images. Rather than reconstructing the complete surface or local details of the target, we emphasize the position, size, orientation, and relationships among individual components.

\subsection{3D Structured Representation}
\subsubsection{Parallel structure}
Recent studies have focused on learning structured 3D models from various modalities, which can be classified into parallel and hierarchical structures. For instance, works such as~\cite{tulsiani2017learning},~\cite{yang2021unsupervised} abstract 3D shapes as several rectangular cuboids, while others~\cite{Paschalidou_2019_CVPR},\cite{liu2022robust},\cite{alaniz2023iterative},~\cite{liu2023marching} abstract them into superquadrics, capable of representing multiple shapes (e.g., cylinders, spheres, cuboids, ellipsoids, etc.) within a single continuous parameter space. Additionally, components in structured 3D models can be represented by meshes~\cite{chen2020bsp},\cite{paschalidou2021neural},~\cite{gao2019sdm}, voxels\cite{wu2019sagnet}, convexes~\cite{deng2020cvxnet},~\cite{yu2025dpa}, and neural implicit representations~\cite{petrov2023anise},\cite{chen2019bae},\cite{chen2024dae},~\cite{liu2024part123},\cite{tertikas2023generating},~\cite{wu2020pq}, which better fit real-world shapes.

\subsubsection{Hierarchical structure}
Although above works are part-aware, they fail to fully consider the relationships between components and the coarse-to-fine granularity of structure, that is, the hierarchy. The reference\cite{wang2011symmetry},\cite{li2017grass} pioneered the use of tree-structured Symmetry Hierarchy (SYMH) to represent and learn the structure of 3D models. The representation has since been applied in optical image single-view 3D reconstruction\cite{niu2018im2struct}, fine-grained point cloud segmentation\cite{yu2019partnet}, and indoor scene generation\cite{li2019grains}. In addition, the binary tree structure has been extended to an N-ary hierarchical structure to better adapt to large datasets\cite{mo2019structurenet}. Other hierarchical learning approaches, such as~\cite{sun2019learning}, propose adaptive hierarchical cube abstractions. In a similar vein, works like~\cite{paschalidou2020learning} represent each component with a superquadric, while~\cite{yang2022dsg} and~\cite{niu2022rim} employ meshes and neural implicit representations, respectively, to model components.

Compared with the aforementioned methods discussed which primarily focus on optical images, SAR images display several discrete scattering centers due to the unique imaging mechanism, which is naturally sensitive to scattering from certain structures. However, no current work has attempted to leverage or reconstruct the 3D structural information embedded in SAR images, which is the basic motivation of this study.


\section{SAR2Struct Overview}
\begin{figure*}[ht]
    \centering
    \includegraphics[width=0.9\textwidth]{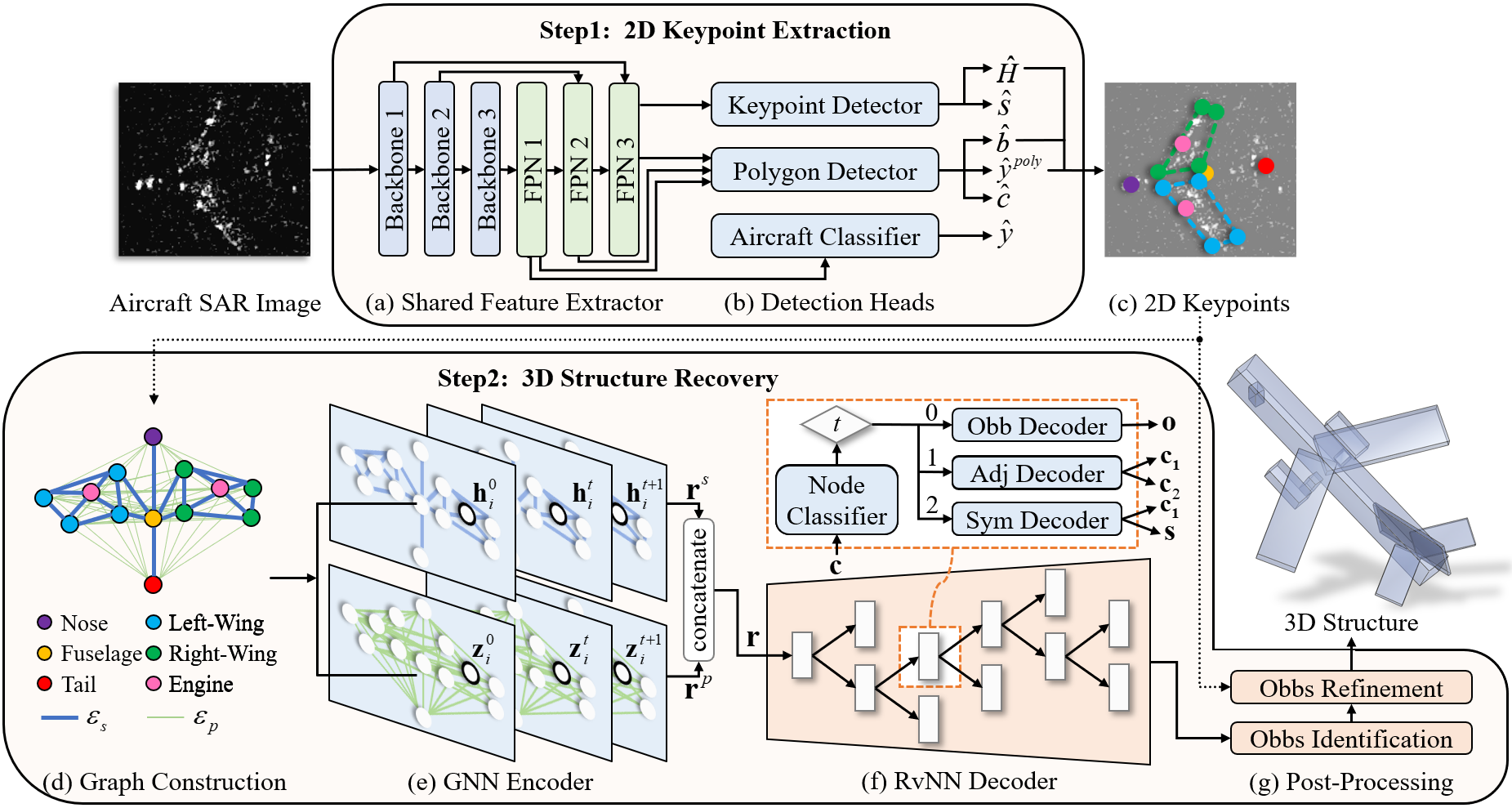}
    \caption{The SAR2Struct process consists of two steps: (1) 2D keypoint extraction and (2) 3D structure recovery. Step 1 includes (a) a shared feature extraction network and (b) detection heads that integrate the predicted heatmap \( \hat{H} \), wing quadrilateral contours \( \hat{b} \), and left-right wing classification \( \hat{y}^{poly} \) into keypoint information for Step 2. In Step 2, (d) a multi-graph is constructed with structure-wise edges (blue) and spatial-wise edges (green), followed by (e) a dual-stream GNN encoding features separately from both edge types. (f) A RvNN progressively decodes node features until all OBBs are obtained. Finally, (g) refine OBBs' parameters through post-processing.}
    \label{fig:pipeline}
\end{figure*}

We introduce SAR2Struct, a two-step algorithmic framework designed to learn the hierarchical structure from a single aircraft SAR image, as shown in Fig.~\ref{fig:pipeline}. In this section, we explain how to represent the hierarchical structure of aircraft and provide an overview of the rationale and approach behind the two-stage algorithm. The neural network models and training strategies used in step 1 and 2 will be described in detail in Sections 4 and 5, separately.

\subsection{Structure Representation}
In this paper, we use SYMH~\cite{wang2011symmetry} to represent the structure of aircraft, including the components and their interrelationships. Each component is abstracted as an oriented bounding box (OBB), encoded with a fixed-length representation vector. The information used to describe the OBB includes its center position (3 dimensions), edge lengths (3 dimensions), and the direction vectors of the three edges (9 dimensions). The third direction vector can be computed as the cross product of the first two, so each OBB is represented by a 12-dimensional code. In addition, two types of relationships between components are defined: adjacency and symmetry.

Fig.~\ref{fig:teaser1}(a) shows the SYMH representation of an aircraft, where each node represents a set of components, and leaf nodes represent individual components. When two nodes correspond to OBBs with an adjacency relationship, their parent node is defined as an adjacent node. When a node corresponds to OBBs that are symmetric to each other, its parent node is defined as a symmetric node, with its sibling node (not shown) storing symmetry information in a 6-dimensional encoding. This paper defines only one type of symmetry for aircraft components: reflectional symmetry (such as the left and right wings), with parameters including the normal vector and center position of the symmetry plane (both 3 dimensions).

\subsection{Two-Step Algorithm}
End-to-end learning of the mapping from SAR images to SYMH is a challenging task. On one hand, there is a lack of paired datasets of target SAR images and 3D models. For available real-world aircraft images, the true 3D shapes cannot be confirmed. An alternative approach could involve generating simulated SAR images from 3D models in ShapeNet\cite{chang2015shapenet}, which would create paired data for supervised training. However, due to the  inherent and difficult-to-model domain gap between simulated and real images, the trained model may be difficult to generalize to real data.

The key issue is how to effectively utilize SAR real-world images and 3D models that do not have direct matching relationships. Our solution is to introduce a physically meaningful intermediate descriptor as a bridge, which can establish matching relationships with both SAR images and 3D shapes, thereby decomposing the original problem into two solvable subproblems. For the aircraft structure recovery task in this paper, we select component keypoints as the intermediate descriptors (Fig.~\ref{fig:teaser2}), deliberately ignoring textures and other features. This choice allows us to adequately represent the dependency between SAR images and 3D structures. Consequently, the structure recovery task is decomposed into two steps: extracting keypoints from aircraft SAR images and learning the mapping from keypoints to SYMH.


\section{2D Keypoint Detection}

In the first step of SAR2Struct, we propose a multi-task learning framework, combining a shared feature extraction network with task-specific decoders, and enhanced by an adaptive multi-task training strategy, to accurately extract component of the aircraft from SAR images.

\subsection{Fine-Grained Annotation of Aircraft SAR Images}
Traditional SAR image aircraft annotation methods often rely on marking aircraft targets with horizontal or rotated rectangular bounding boxes, focusing primarily on category labeling. However, such methods fail to meet the demand for fine-grained interpretation of aircraft structural information. To precisely locate the key components of aircraft targets, this study adopts a fine-grained annotation strategy consisting of the following three aspects as depicted in Fig.~\ref{fig:pipeline}(c):
\subsubsection{Category Annotation}
Aircraft targets in the images are categorized in detail to indicate their specific types.
\subsubsection{Keypoint Annotation} Precise annotations of the coordinates for three key locations, including nose, fuselage center, and tail, are provided for each aircraft target, capturing the spatial distribution of the aircraft body and its orientation. For aircraft  with prominent engine features, engine positions are also separately annotated.
\subsubsection{Polygonal Bounding Box Annotation} For the left and right wings of the aircraft, irregular quadrilateral bounding boxes are employed, which can be used to distinguish between the left and right wing categories.

\subsection{Keypoint Detection Network}
Based on the fine-grained aircraft component annotation strategy proposed above, we design a network to jointly learn the following three tasks: aircraft classification, keypoint detection for aircraft components, and irregular polygonal bounding box detection of wings. We propose a multi-task learning framework that shares parameters to improve task performance while enhancing training efficiency. The framework is shown in Fig.~\ref{fig:pipeline}(a)-(b) and consists of the following two key parts:

\subsubsection{Shared Feature Extraction}  
This module utilizes a pre-trained ResNet as the backbone, combined with a three-layer Feature Pyramid Network (FPN) for feature extraction. The shared feature extraction network is designed to extract general feature representations from the input SAR images, leveraging the shared characteristics across tasks and eliminating redundant feature computation.

\subsubsection{Detection Heads}  
To achieve multi-task interpretation, task-specific detection decoders are designed on top of the shared feature extraction network:

\textbf{Classifier:} This classifier connects to shallow features from the FPN, capturing global information from the image and predicting a multi-class probability vector to accomplish the classification task of aircraft targets. The corresponding loss function can be formulated as:
\begin{equation}
    \mathcal{L}_{aircls} = -\frac{1}{N_b}\sum_{i=1}^{N_b} y_i \log(\hat{y}_i),
\end{equation}
where $y$ and $\hat{y}$ represent the true and predicted labels of the aircraft image, respectively, and $N_b$ is the batch size.

\begin{figure}[hbt]
	\centering
	\includegraphics[width=\linewidth]{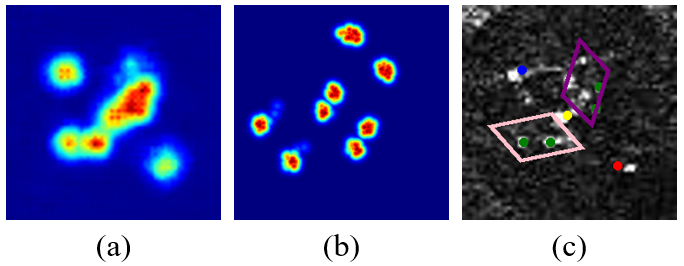}
	\caption{Visualization of the detection head outputs. (a) The heatmap $\hat{H}$ output by the keypoint detector; (b) The wing $\hat{b}$ output by the polygon detector; (c) The keypoint information extracted in Step 1.}\label{fig:heatmap}
\end{figure}

\textbf{Keypoint Detector:} This module is connected to deep feature maps of the FPN for fine-grained features. Following heatmap-based keypoint detection methods~\cite{xiao2018simple}, the module first predicts a globally distributed heatmap (Fig.~\ref{fig:heatmap}(a)), and then identifies the coordinates and confidence scores of keypoints based on the heatmap distribution. To accommodate aircraft with variable engine configurations or missing engine labels, the module filters keypoints based on score thresholds derived from the heatmap, improving its flexibility. The corresponding loss function is:
\begin{equation}
    \mathcal{L}_{kp} = w_{heatmap} \cdot {L}_{heatmap} + w_{score} \cdot {L}_{score},
\end{equation}
where
\begin{equation}
    {L}_{heatmap}(\hat{H}, H) = \frac{1}{N_b} \sum_{i=1}^{N_b} \|\hat{H}_i - H_i\|^2,
\end{equation}
\begin{equation}
\begin{aligned}
    {L}_{score}(v, \hat{s}) &= -\frac{1}{N_b} \sum_{i=1}^{N_b} \left[ v_i \log(\sigma(\hat{s}_i)) \right. \\
    &\quad + \left. (1 - v_i) \log(1 - \sigma(\hat{s}_i)) \right],
\end{aligned}
\end{equation}
where $H$ and $\hat{H}$ represent the ground truth and predicted heatmaps, $v$ and $\hat{s}$ represent the ground truth visibility and predicted confidence of the keypoints, and $w_{heatmap}$ and $w_{score}$ are the corresponding loss weights.

\textbf{Polygon Detector:} This module is designed to predict the contours of aircraft wings by leveraging multi-layer feature maps from the FPN. It locates polygons and predicts wing categories along with confidence scores. While rectangular bounding boxes require only two diagonal coordinates for positioning and straightforward IoU calculations, irregular polygons rely on four vertex coordinates, which increase the computational complexity of IoU. To address this challenge, we introduce a parallel computation strategy for irregular polygonal IoU~\cite{gloudemans2023polygon}, enhancing the computation efficiency. The corresponding loss function can be formulated as:
\begin{equation}
    \mathcal{L}_{poly} = w_{loc} \cdot {L}_{loc} + w_{ploycls} \cdot {L}_{ploycls} + w_{conf} \cdot {L}_{conf},
\end{equation}
where

\begin{equation}
    {L}_{loc}(b, \hat{b}) =
    \begin{cases}
        \frac{1}{2N_b} \sum\limits_{i=1}^{N_b} \|b_i - \hat{b}_i\|^2,\text{ if } \|b_i - \hat{b}_i\| < 1, \\
        \frac{1}{N_b} \sum\limits_{i=1}^{N_b} \left(\|\hat{b}_i - b_i\| - 0.5\right), \text{otherwise}.
    \end{cases}
\end{equation}
\begin{equation}
    {L}_{polycls}(y^{{poly}}, \hat{y}^{{poly}}) = -\frac{1}{N_b} \sum_{i=1}^{N_b} y^{{poly}}_i \log(\hat{y}^{{poly}}_i),
\end{equation}
\begin{equation}
\begin{aligned}
    {L}_{{conf}}&(\hat{c}, {IoU}, {pos}, {neg}) = - \frac{1}{N_{{pos}}} \sum_{i \in {pos}} 
    [ {IoU}_i\\
    &\cdot \log(\sigma(\hat{c}_i)) + (1 - {IoU}_i) \cdot \log(1 - \sigma(\hat{c}_i))] \\
    & - \frac{\alpha}{N_{{neg}}} \sum_{i \in {neg}} \log(1 - \sigma(\hat{c}_i)),
\end{aligned}
\end{equation}

\noindent where $b$ and $\hat{b}$ represent the ground truth and predicted polygons of the wings, $y^{poly}$ and $\hat{y}^{poly}$ represent the ground truth and predicted polygon labels, $\hat{c}$, ${IoU}$, ${pos}$ and ${neg}$ refer to the predicted confidence scores, IoU values, indices of positive and negative samples, respectively. $w_{loc}$, $w_{polycls}$ and $w_{conf}$ are the corresponding loss weights, ${\alpha}$ is oefficient ratio of positive and negative samples.

\subsection{Adaptive Multi-Task Training Strategy}
In multi-task learning, varying loss magnitudes and task influences can cause gradient imbalances, particularly in shared parameters, leading to conflicts and reduced model performance. A common approach combines task losses with fixed weights, which in this aircraft component interpretation task, can be formulated as:
\begin{equation}
\mathcal{L}_{total} = w_{aircls} \mathcal{L}_{aircls} + w_{kp} \mathcal{L}_{kp} + w_{poly} \mathcal{L}_{poly},
\label{loss1}
\end{equation}
\noindent where $w_{aircls}$, $w_{kp}$, and $w_{poly}$ are the weights for the respective loss terms. While this static weighting strategy is straightforward and easy to implement, it lacks flexibility and often results in an imbalance in task prioritization. This paper introduces an adaptive loss training strategy that balances training efficiency and model performance by combining dynamic task weighting and gradient optimization. The approach has two stages:
\subsubsection{Coarse-granularity gradient optimization stage} 
Dynamic Weight Averaging (DWA)~\cite{liu2019end} is employed to adjust task weights based on historical loss variations, assigning higher weights to tasks with greater changes. This balances task learning rates and prevents any single task from dominating shared parameter optimization.
\subsubsection{Fine-granularity gradient optimization stage}
Task priority~\cite{jeong2024quantifying} is used to classify shared parameter convolutional layers at the channel level, identifying the dominant task for each channel. Gradients are then fine-tuned at the channel level, enabling targeted optimization of shared parameters.


\section{3D Structure Recovery}

In the second step of SAR2Struct, we learn mapping from the detected component keypoints to the 3D structure representation. This step requires solving two key problems: how to effectively extract depth features from 2D coordinates of the component keypoints, and how to recover hierarchical structures from these depth features.

To address the first problem, we construct an undirected multi-graph to model the spatial and structural relationships between keypoints, and then develop a dual-stream GNN to fully extract features from the multi-graph~\cite{jiang2021recognizing},~\cite{dou2024hierarchical}. After multiple rounds of node feature aggregation and updates, the entire graph is represented as a $d$-dimensional feature code (with $d$=80 in this paper), corresponding to the root node in SYMH. Next, using a RvNN~\cite{li2017grass}, the root code is recursively decoded into features of component sets in the 3D model, and finally features of each leaf node are further decoded into OBB parameters, thereby obtaining the SYMH representation.

\subsection{Graph Construction}
\textbf{Nodes.} The keypoint information includes 2D coordinates of the nose, fuselage center, tail, engines, and four vertices of bounding quadrilateral for the left and right wings. Each component keypoint is constructed as a node in the graph. For a keypoint $p$, its corresponding node attributes $x$ consist of the point coordinates $(p^x, p^y)$ and the component type \( y^{node} \):
\begin{equation}
    x = \text{concat}(p^x, p^y, y^{node}),  p \in \mathcal{P},
\end{equation}
where $\mathcal{P}$ denotes the set of all keypoints in the sample.

\textbf{Edges.} A multi-graph consisting of two sets of edges is constructed: structure-wise edges and spatial-wise edges, which capture keypoint features and their dependencies from different perspectives. As shown by the blue lines in Fig.~\ref{fig:pipeline}(d), we connect the following nodes to form structure-wise edges $\mathcal{E}_s$: nose-fuselage, fuselage-tail, fuselage-left wing inner vertices, fuselage-right wing inner vertices, left engine- four vertices of the left wing, and right engine-four vertices of the right wing, which can be represented as:
\begin{equation}
\mathcal{E}_s = \left\{ (p_i, p_j) : p_i, p_j \in \mathbb{S} \right\},
\end{equation}
where $\mathbb{S}$ represents the set of node pairs $(p_i, p_j)$ connected by structure-wise edges. These edges model the relationships based on structural semantics of the aircraft, while ignoring spatial proximity. To better capture the spatial dependencies between nodes, as shown by the green lines in Fig.~\ref{fig:pipeline}(d), spatial-wise edges $\mathcal{E}_p$ is introduced, which are defined as dense connections between all nodes in the sample:
\begin{equation}
\mathcal{E}_p = \left\{ (p_i, p_j) : p_i, p_j \in \mathbb{P} \right\},
\end{equation}
where $\mathbb{P}$ represents the set of node pairs $(p_i, p_j)$ connected by spatial-wise edges.

\subsection{Graph Neural Encoder}
A dual-stream GNN architecture is adopted, as shown in Fig.~\ref{fig:pipeline}(e), with two branches updating node representations based on structure-wise and spatial-wise edges, respectively. After a specified number of updates, the features obtained from both branches are fused.

\subsubsection{Structure-wise Stream}  
For the graph constructed with structure-wise edges, when computing the message for node $i$ at timestep $t+1$, we consider the concatenation of its node feature $\mathbf{h}_i^t$ and difference between its feature and that of its neighboring nodes $\mathbf{h}_j^{t}$ from the previous timestep. Based on average aggregation of messages, the update rule for node $i$ at timestep $t+1$ is as follows:
\begin{equation}
    \mathbf{h}_i^{t+1} = f^l\left(\mathbf{h}_i^t\right) + \frac{1}{|\mathcal{N}_i^s|} \sum_{j \in \mathcal{N}_i^s} f^s\left(\text{concat}(\mathbf{h}_i^t, \mathbf{h}_j^t - \mathbf{h}_i^t)\right), \label{equ:gnn}
\end{equation}
where $\mathcal{N}_i^s$ denotes the set of neighbors of node $i$ based on structure-wise edges. The function $f^s$ is a multi-layer perceptron comprising two fully connected layers, ReLU activation, and batch normalization. The function $f^l$ is a linear layer.

\subsubsection{Spatial-wise Stream}  
For the dense graph constructed with spatial-wise edges, when computing the message for node $i$ at timestep $t+1$, we only consider the node features $\mathbf{z}_j^t$ of its neighboring nodes $j$ at timestep $t$, using average aggregation. The update rule for node $i$ at timestep $t+1$ is as follows:
\begin{equation}
    \mathbf{z}_i^{t+1} = \frac{1}{|\mathcal{N}_i^{p}|} \sum_{j \in \mathcal{N}_i^{p} \cup \{i\}} f^p(\mathbf{z}_j^t),
\end{equation}
where $\mathcal{N}_i^{p}$ denotes set of all other nodes in the graph. The function $f^p$ is a fully connected layer with ReLU activation and batch normalization.

\subsubsection{Feature Fusion}  
After updating, features from two branches are fused. First, node features from each branch at every timestep are concatenated to form $\mathbf{r}_i^s$ and $\mathbf{r}_i^p$, and then mapped through respective perceptrons $f^1$ and $f^2$. The structure-wise features are aggregated by taking the maximum value of each node’s features, resulting in a global graph feature of dimension $d/2$, while the spatial-wise features are aggregated by averaging each node’s features, resulting in another global feature of dimension $d/2$. These two features are then concatenated to form the final $d$-dim feature code $\mathbf{r}$:
\begin{equation}
\begin{aligned}
\mathbf{r} = \text{concat} ( & \max_{i \in \mathcal{N}}( \text{concat}( f^1(\mathbf{r}_i^s), \mathbf{r}_i^s ) ), \\
& \frac{1}{|\mathcal{N}|} \sum_{i \in \mathcal{N}} ( \text{concat}( f^2(\mathbf{r}_i^p), \mathbf{r}_i^p ) ) ),
\end{aligned}
\end{equation}
where
\begin{equation}
\mathbf{r}^s_i = \text{concat}( \mathbf{h}_i^0, \mathbf{h}_i^1, \dots, \mathbf{h}_i^T ),
\end{equation}
\begin{equation}
\mathbf{r}^p_i = \text{concat}( \mathbf{z}_i^0, \mathbf{z}_i^1, \dots, \mathbf{z}_i^T ),
\end{equation}
where $\mathcal{N}$ represents set of all nodes, and $T$ denotes the total number of update steps. \( f^1 \) and \( f^2 \) are both composed of a fully connected layer, ReLU activation function, and batch normalization, with network structure being the same but without shared weights.

\subsection{Recursive Neural Decoder}
The feature encoding output $\mathbf{r}$ by dual-stream GNN is fed into downstream decoder to learn SYMH representation of the aircraft, as shown in Fig.~\ref{fig:pipeline}(f). According to the definition of SYMH tree in Section III, the nodes are categorized into adjacency nodes, symmetry nodes, and OBB nodes (i.e., leaf nodes). This section describes how the RvNN Decoder is constructed and how to design the node classifier and decoders for different types of nodes.

\begin{figure}[hbt]
	\centering
	\includegraphics[width=0.8\linewidth]{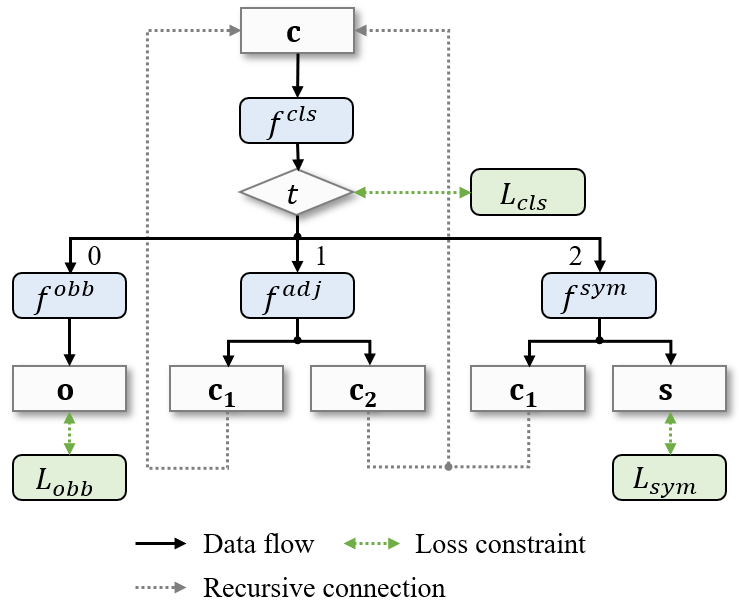}
	\caption{Schematic of recursive decoding.}\label{fig:rvnn}
\end{figure}

\subsubsection{Node Classifier}
This classifier predicts category of each node based on its feature encoding \( \mathbf{c} \), where \( t = 0, 1, 2 \) corresponds to OBB, adjacency, and symmetry nodes, respectively. Based on the prediction results, \( \mathbf{c} \) is fed into the corresponding decoder:
\begin{equation}
    t = f^{cls}(\mathbf{c}), \quad f^{cls}: \mathbb{R}^d \to \{0, 1, 2\}.
\end{equation}

\subsubsection{Adjacency Decoder}
This decoder decodes a \( d \)-dim feature of parent node into two child nodes \( \mathbf{c}_1 \) and \( \mathbf{c}_2 \):
\begin{equation}
    (\mathbf{c}_1, \mathbf{c}_2) = f^{adj}(\mathbf{c}), \quad f^{adj}: \mathbb{R}^d \to \mathbb{R}^d \times \mathbb{R}^d.
\end{equation}

\subsubsection{Symmetry Decoder}
This decoder decodes a \( d \)-dim feature of the parent node into a child node \( \mathbf{c} \) and a 6-dim symmetry parameter \( \mathbf{s} \), whose meaning has been described in Section III:
\begin{equation}
    (\mathbf{c}, \mathbf{s}) = f^{sym}(\mathbf{c}), \quad f^{sym}: \mathbb{R}^d \to \mathbb{R}^d \times \mathbb{R}^6.
\end{equation}

\subsubsection{OBB Decoder}
This decoder decodes a leaf node described by a \( d \)-dim feature into 12-dim OBB parameters \( \mathbf{o} \), whose meaning has been illustrated in Section III:
\begin{equation}
    (\mathbf{o}) = f^{obb}(\mathbf{c}), \quad f^{obb}: \mathbb{R}^d \to \mathbb{R}^{12}.
\end{equation}

Fig.~\ref{fig:rvnn} illustrates the recursive decoding process. The mean squared error between all decoded \( \mathbf{o} \) and $\mathbf{s}$ and their corresponding ground truth is computed. At the same time, cross-entropy loss is used to guide the classifier’s training, ensuring that the SYMH tree unfolds correctly:
\begin{equation}
    L = \lambda_{\text{cls}} \cdot \text{CE}(t, \hat{t}) + \lambda_{\text{sym}} \cdot \text{MSE}(\mathbf{s}, \mathbf{\hat{s}}) + \lambda_{\text{obb}} \cdot \text{MSE}(\mathbf{o}, \mathbf{\hat{o}}),
\end{equation}
where \( w_1, w_2, w_3 \) are weight hyperparameters used to balance the contribution of each loss term. During training, ground truth for hierarchical structure of the keypoints is required. During testing, however, ground truth are not needed and the network can directly infer hierarchical structure from the input keypoints.

In the testing phase, we introduce a rule-based post-processing mechanism that refines OBBs generated by the network, leveraging input 2D keypoints, as shown in Fig.~\ref{fig:pipeline}(g). The OBB category is identified based on predefined rules and mapped to the corresponding 2D component first. During fuselage refinement phase, the center position, length, and direction are calculated using keypoints, then adjust the OBB alignment. Similarly, for wings we compute their center and dimensions, adjusting the long and short edges based on directional vectors. Regarding the engines, the Hungarian algorithm~\cite{kuhn1955hungarian} is employed to match the predicted OBB center with the actual engine keypoint coordinates, with OBB parameters being updated accordingly. This post-processing mechanism progressively refines OBB parameters output by the network, improving recovery accuracy.


\section{Experiments}

\subsection{Data Synthesis}
The keypoint detection task utilized a dataset derived from the GF-3 satellite, comprising a total of 5156 image slices of 21 aircraft types, with a resolution of 1 meter. The dataset was split into training and validation sets at a ratio of 4:1 of each aircraft type, resulting in 4116 training images and 1040 validation images.


In the structure recovery phase, training data pairs were synthesized using aircraft models from the ShapeNet dataset~\cite{chang2015shapenet}, including SYMH and 2D keypoints. For SYMH synthesis, we use fine-grained segmentation data of aircraft processed in~\cite{sun2023semi}. OBBs are fitted to each segment. The adjacency and symmetry relationships are then detected, resulting in the component relationship graph of the aircraft. Finally, the graph is converted into SYMH format using several priority rules and iterative graph contraction algorithm~\cite{wang2011symmetry}.

\begin{figure}[t]
	\centering
	\includegraphics[width=0.65\linewidth]{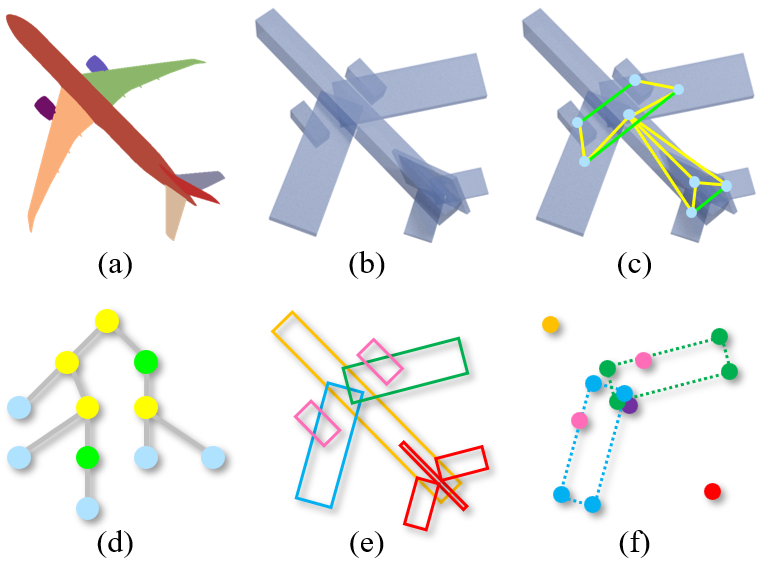}
	\caption{Dataset synthesis process for step 2. (a) Fine-grained segmented data. (b) OBB fitting. (c) Adjacency (yellow lines) and symmetry (green lines) detection. (d) Graph contraction to generate a SYMH. (e) Projection of OBBs onto a 2D plane followed by MRB fitting. (f) Extraction and alignment of 2D keypoints.}\label{fig:data2}
\end{figure}

To ensure that the synthesized 2D keypoints have consistent representation with the labeled data used in Step 1, we first project the 3D model onto 2D plane and fit a minimum bounding rectangle (MBR) to each part. Then, classification rules are applied based on the aircraft's shape characteristics to determine the component type for each rectangle. After information filtering and format alignment, the final training pairs consist of SYMH and 2D keypoints as shown in Fig.~\ref{fig:data2}(d) and Fig.~\ref{fig:data2}(f). A total of 1563 samples were synthesized, with 1243 for training, 160 for validation, and 160 for testing.

\subsection{Keypoint Detection and Recognition}

\subsubsection{Experimental Setup}  
The experiments were conducted on an RTX 3090 GPU, with all images resized to 256×256 pixels. Through preliminary testing, the required training epochs for each sub-task were determined as 1000, 200, and 2000. To unify the training process, the number of training epochs was set to 2,000 across all tasks, with an early stopping strategy implemented to improve efficiency.

The learning rate schedule was defined as follows: the initial learning rate was set to $5 \times 10^{-5}$, and the batch size was set to 256 (reduced to 128 during the second stage of the multi-task adaptive learning strategy). For the first 50 epochs, a linear warm-up strategy was used to gradually increase the learning rate to $5 \times 10^{-4}$. Subsequently, a cosine annealing learning rate schedule was applied, with an initial cycle length of $T_0 = 10$, a growth factor $T_{\text{mult}} = 2$, and a minimum learning rate $\eta_{\text{min}} = 5 \times 10^{-6}$. The Adam optimizer was employed.


\begin{table*}[ht]
    \centering
    \caption{Performance comparison of experimental results in step 1}
    \label{results1}
    \begin{tabular}{lccccccc}
        \toprule
        Methods & Loc Error $\downarrow$ & Ang Error $\downarrow$ & $AP_{0.5}$ $\uparrow$ & $AR_{0.5}$ $\uparrow$ & OA $\uparrow$ & $mAP_{0.5}$ $\uparrow$ & F1 $\uparrow$ \\
        \midrule
        Independent & 11.8090 & 21.3604 & 0.7800 & 0.7815 & 0.8067 & 0.6005 & 0.7292 \\
        Uniform & 11.6272 & 20.2903 & 0.7722 & 0.7812 & 0.8221 & 0.6163 & 0.7349 \\
        Ours & \textbf{11.3840} & \textbf{19.1626} & \textbf{0.8099} & \textbf{0.7823} & \textbf{0.8538} & \textbf{0.6497} & \textbf{0.7651} \\
        \bottomrule
    \end{tabular}
\end{table*}

To evaluate the effectiveness of the proposed model framework, three comparative experiments were designed:
\begin{itemize}
    \item \textbf{Independent:} The model structure remained fixed, and only the loss function of a single task was optimized and backpropagated. Detection heads for other tasks were not included in the training process.
    \item \textbf{Uniform:} Using Equation~\ref{loss1}, all task loss weights were set to 1, meaning the loss functions of all tasks were summed directly, and the gradients were backpropagated without adjustment.
    \item \textbf{Ours:} The proposed multi-task adaptive learning strategy was employed, dynamically adjusting the gradient optimization during training.
\end{itemize}

\subsubsection{Evaluation Metrics}
To comprehensively verify the effectiveness of the proposed model framework, we select common evaluation metrics for each sub-task as follows:
\begin{itemize}
    \item \textbf{Classification task:} Overall accuracy (OA) was used as the evaluation criterion.
    \item \textbf{Keypoint detection task:} Location Error: The average distance error between predicted and ground truth keypoints. Angle Error: The aircraft orientation angle was predicted using the least squares method based on three keypoints, and the error was measured as the difference between predicted and actual angles. Average Precision and Recall: Object Keypoint Similarity (OKS)~\cite{lin2014microsoft} was used with a confidence score 0.5 filtering standard to calculate the Average Precision (AP0.5) and Average Recall (AR0.5).
    \item \textbf{Wing bounding box detection task:} F1-score: Combines precision and recall to evaluate detection accuracy and completeness. mAP0.5: Mean Average Precision at an IoU threshold of 0.5, where higher values reflect better bounding box detection performance.
\end{itemize}

\begin{figure}[t]
	\centering
	\includegraphics[width=\linewidth]{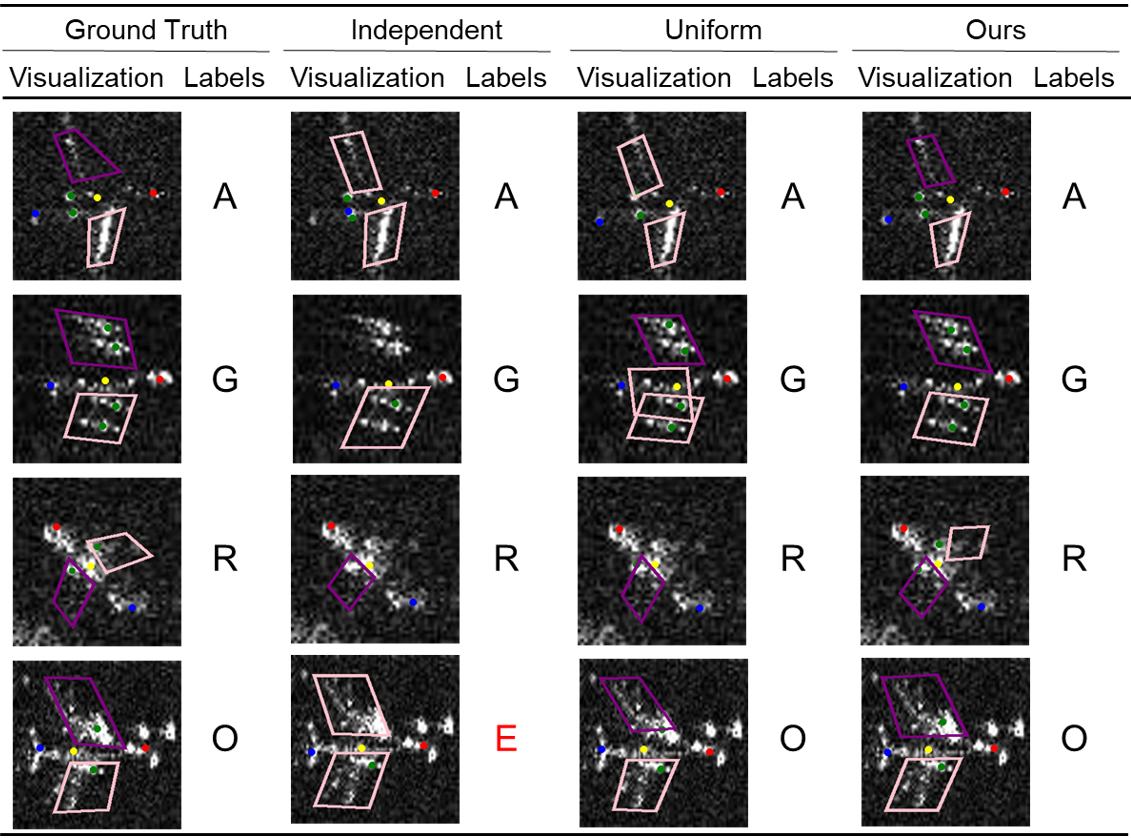}
	\caption{Visualization of predicted categories and key components. The different colors represent the detected components, with A, G, R, and O indicating different aircraft categories.}\label{fig:results1}
\end{figure}

\subsubsection{Experimental Results}
The experimental results, summarized in Table~\ref{results1}, demonstrate that the uniform multi-task learning strategy outperforms independent training of each sub-task. Moreover, with the introduction of the proposed adaptive multi-task learning strategy, all performance metrics reached their optimal levels.

\begin{figure*}[t]
    \centering
    \includegraphics[width=0.85\textwidth]{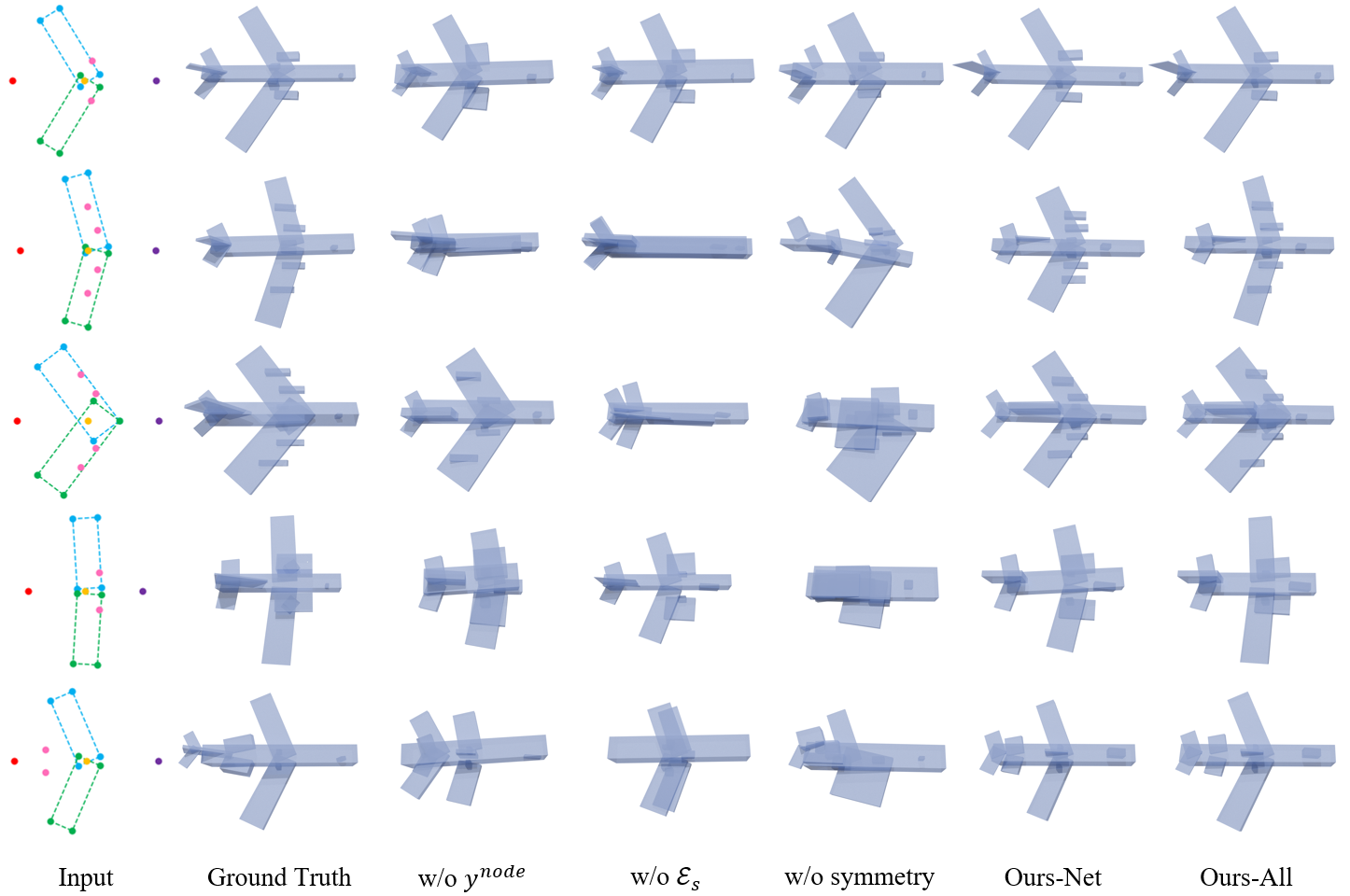}
    \caption{Visualization and comparison. Compared to the baseline methods, Ours-Net generates more accurate OBBs in position and shape, while Ours-All further refines the predictions geometrically to improve accuracy.}
    \label{fig:results2}
\end{figure*}

Fig.~\ref{fig:results1} visualizes some results, with the letters on the right indicating predicted target categories. Common issues include missed, over-detected, or misclassified wing bounding boxes, missing engine keypoints, positional deviations in body keypoints, and occasional classification errors. Even under optimal conditions, challenges remain for tasks such as accurately localizing irregular wing bounding boxes and predicting precise keypoint coordinates, leaving room for improvement.

\subsection{Structure Recovery with Simulated Data}
\subsubsection{Experimental Setup}  
The network is performed on a desktop computer equipped with an Intel Core i5-13400F CPU and a GeForce RTX 4060 GPU (24GB memory). The optimizer used is Adam, with a learning rate of \( 1 \times 10^{-4} \) and a weight decay of \( 1 \times 10^{-5} \). The learning rate is scheduled using a StepLR policy with a step size of 50 and a decay factor of 0.8. The training process consists of 200 epochs, and batch size is set to 64.


\subsubsection{Evaluation Metrics}  


We refer to the Hausdorff Error \( E_H \)~\cite{niu2018im2struct} to measure the geometric shape difference between predicted OBBs and ground truth. While the Hausdorff Error directly measures the difference between the boundary vertices of the OBBs and is sensitive to deviations, we also calculate the 95\% Hausdorff Error to mitigate the impact of rare outliers. In this case, the minimum distances between all pairs of points are computed, and the 95th percentile of these distances is taken.

In addition, IoU is used to assess the overall similarity between the predicted OBB sets $S_i$ and ground truth $S_i^{gt}$. The internal spaces of the OBBs are voxelized and the IoU is computed between voxel sets \( V_i \) and \( V_i^{gt} \):

\begin{equation}
IoU = \frac{1}{N} \sum_{i=1}^N \frac{|V_i \cap V_i^{gt}|}{|V_i \cup V_i^{gt}|}.
\end{equation}

Apart from these geometric accuracy metrics for OBBs, we introduce the Subtree Matching Score (SMS) to evaluate the topological consistency between predicted and ground truth hierarchical structures. Since tree structures are represented as pre-order traversal node lists, the consistency can be easily checked by comparing nodes and recursively checking subtrees. Starting from the root node, we count the number of nodes with exactly matching subtree structures in predicted and ground truth trees ${M_{subtree}}$, and define the SMS as:

\begin{equation}
SMS = \frac{M_{subtree}}{\max(N_{node}^{pred}, N_{node}^{gt})},
\end{equation}
where \( N_{node}^{pred} \) and \( N_{node}^{gt} \) represent number of nodes in predicted and true trees, respectively. This metric effectively quantifies the topological consistency of structure predictions.

\subsubsection{Structure Recovery Results}
For the quantitative evaluation of step 2 algorithm, it was compared against five baselines, which can be regarded as ablation versions of our method. We focus on the necessity of including component category $y^{node}$ in node attributes during graph construction, the performance of the two sets of edges and their respective branches in the dual-stream GNN, and the role of symmetry in hierarchical structure representation. This led to the design of the five benchmark models in Table~\ref{results2}.

From Table~\ref{results2} and Fig.~\ref{fig:results2}, it can be seen that our complete model achieves the best performance across all metrics with the assistance of all design elements. Moreover, we find that: (1) Removing $h_j^t - h_i^t$ from the structure-wise stream results in the most significant performance drop in OBB reconstruction in graph construction and GNN model, indicating that considering feature differences between neighboring nodes during aggregation enhances the encoding performance of the graph neural encoder. (2) Removing the component type information $y^{node}$ has the least impact on OBB reconstruction but significantly affects SYMH tree reconstruction (SMS), as the relationship between tree structure and component types is highly correlated. (3) Removing either structure-wise edges $\mathcal{E}_s$ or spatial-wise edges $\mathcal{E}_p$ leads to a performance drop, as both edges encode relationships between components from different perspectives.

\begin{figure}[t]
	\centering
	\includegraphics[width=\linewidth]{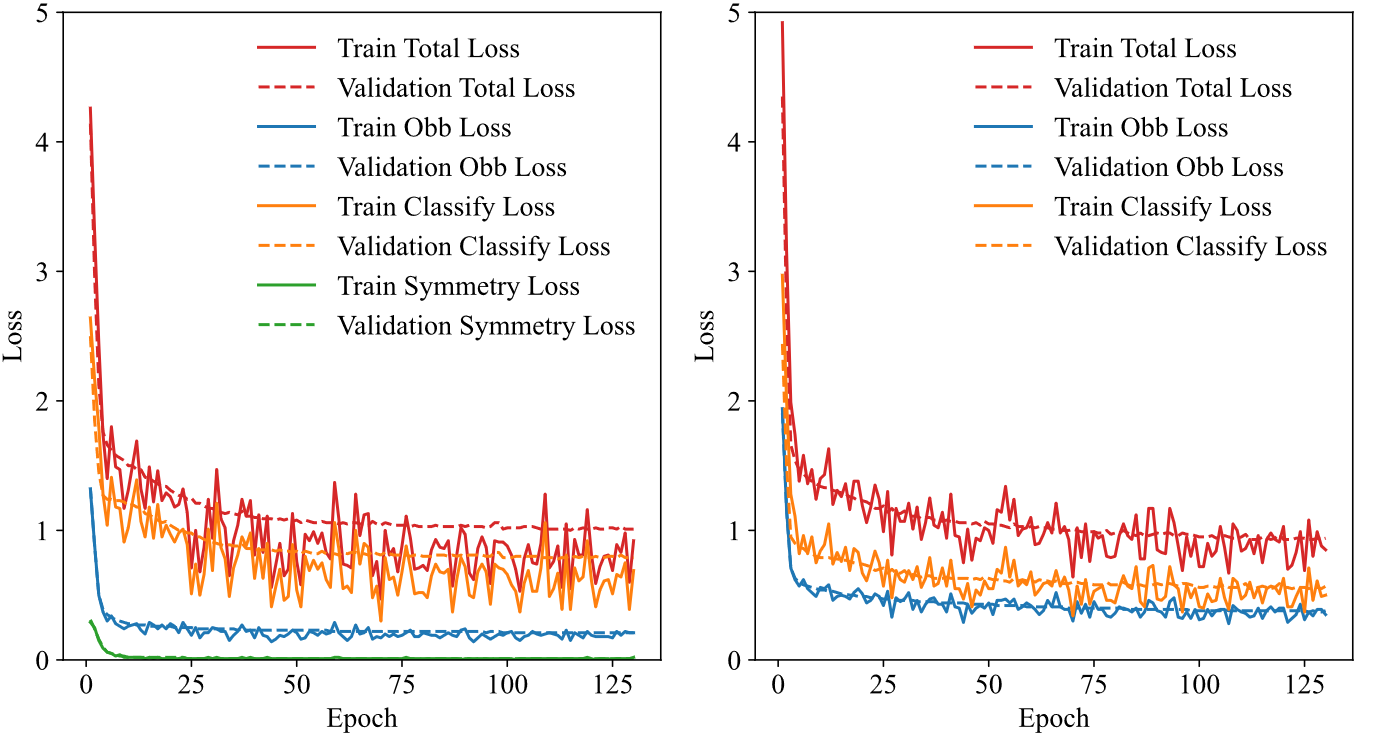}
	\caption{Training process loss curves. Left: Representation with symmetry consideration; Right: Representation without symmetry consideration.}\label{fig:loss}
\end{figure}

Our experimental results show that removing symmetry from the representation significantly lowers the accuracy of aircraft structure recovery. This aligns with our expectations, as the 3D shape of aircraft has high symmetry, and losing this crucial prior constraint results in a noticeable drop in reconstruction IoU. As seen in the fifth column of Fig.~\ref{fig:results2}, the symmetry of left-right wings, left-right tail wings, and left-right engines is lost. Fig.~\ref{fig:loss} shows the training process with and without symmetry representation. Without symmetry constraint, the representation becomes redundant, and the node classification is simplified into a binary classification task, reducing the classification loss (yellow line). However, the number of OBB parameters to regress increases, making the reconstruction loss (blue line) higher. Adjusting complexity of node classifier and OBB decoder may improve this, but it is beyond the scope of our current study.

\begin{table}[h!]
    \centering
    \caption{Comparison of our network and full model with five baseline models.}
    \begin{tabular}{lcccccc}
        \toprule
        Methods & \boldmath{$E_H$ ↓} & \boldmath{$95\% E_H$ ↓} & \boldmath{IoU} ↑ & \boldmath{SMS} ↑ \\
        \midrule
        w/o $y^{node}$         & 0.1623 & 0.1603 & 0.5423 & 0.8031 \\
        w/o $\epsilon_s$       & 0.1696 & 0.1675 & 0.5272 & 0.8127 \\
        w/o $\epsilon_p$       & 0.1660 & 0.1639 & 0.5365 & 0.8169 \\
        w/o $h_j^t - h_i^t$    & 0.1701 & 0.1681 & 0.5205 & 0.8123 \\
        w/o symmetry           & 0.2082 & 0.2056 & 0.4685 & 0.8256 \\
        Ours-Net               & 0.1537 & 0.1518 & 0.5472 & \textbf{0.8284} \\
        Ours-All               & \textbf{0.1224} & \textbf{0.1203} & \textbf{0.6300} & \textbf{0.8284} \\
        \bottomrule
        \label{results2}
    \end{tabular}
\end{table}

\begin{figure}[ht]
	\centering
	\includegraphics[width=0.9\linewidth]{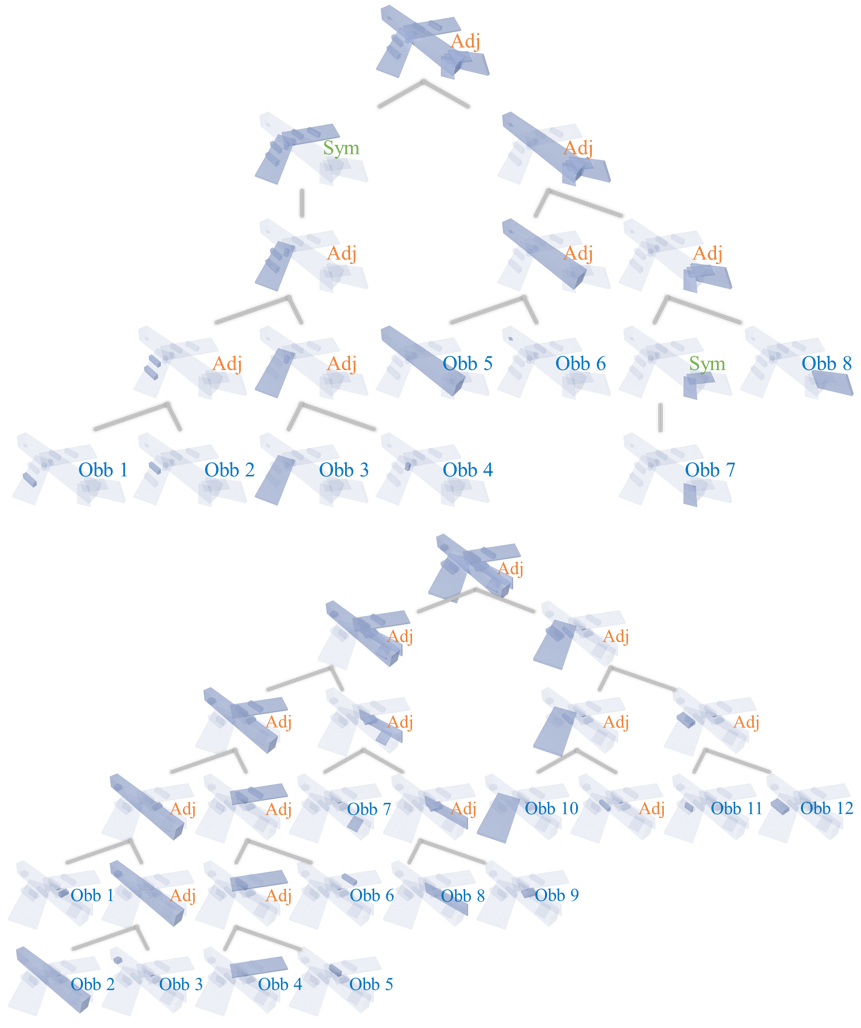}
	\caption{Hierarchical structure of aircraft recovered from the same input. Top: When symmetry is considered in the representation, 8 different-shaped OBBs are required. Bottom: When symmetry is not considered, 12 different-shaped OBBs are needed.}\label{fig:sym}
\end{figure}

Fig.~\ref{fig:sym} shows the reconstructed hierarchical trees with and without symmetry representation. When symmetry is considered, the generated tree structure is more concise and exhibits better semantic consistency in subtree structures. Without symmetry, the tree complexity increases significantly, with redundant nodes and branches, and similar components may be incorrectly assigned to different branches, weakening the model’s reconstruction ability.

\subsection{Structure Recovery with Real SAR Images}

In this section, the entire two-step algorithm is tested using GF-3 satellite data. The data includes real imaging of various aircraft models under different viewing angles and conditions, providing challenging test scenarios to assess the generalization performance of the algorithm.

\begin{figure*}[t]
	\centering
	\includegraphics[width=0.9\linewidth]{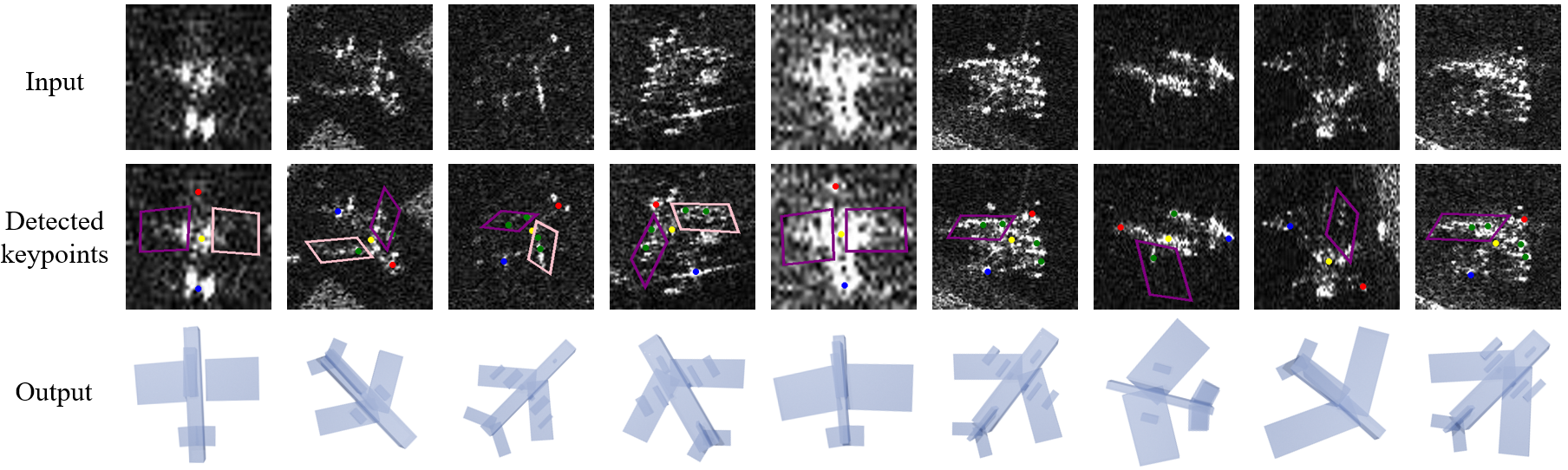}
	\caption{Structure recovery results of real-world aircraft SAR images. The third row shows the visualization after adjusting the orientation manually.}\label{fig:real1}
\end{figure*}

Fig.~\ref{fig:real1} shows the experimental results of SAR2Struct on a few test samples. The first row displays the input SAR images, the second row shows the 2D keypoint information generated by Step 1 of the algorithm, and the third row shows the 3D structure recovered by Step 2 based on the keypoints. As seen in the first 4 samples, Step 1 successfully captures the keypoint information needed by Step 2, providing a reliable foundation for subsequent aircraft structure reconstruction. While for the last 5 samples, due to the complexity of imaging features, Step 1's keypoint detection performance decreases. However, by leveraging the strong prior knowledge of aircraft's high symmetry, the missing information is effectively compensated for, and reasonable structures are still managed to be reconstructed.

\begin{figure}[ht]
	\centering
	\includegraphics[width=0.6\linewidth]{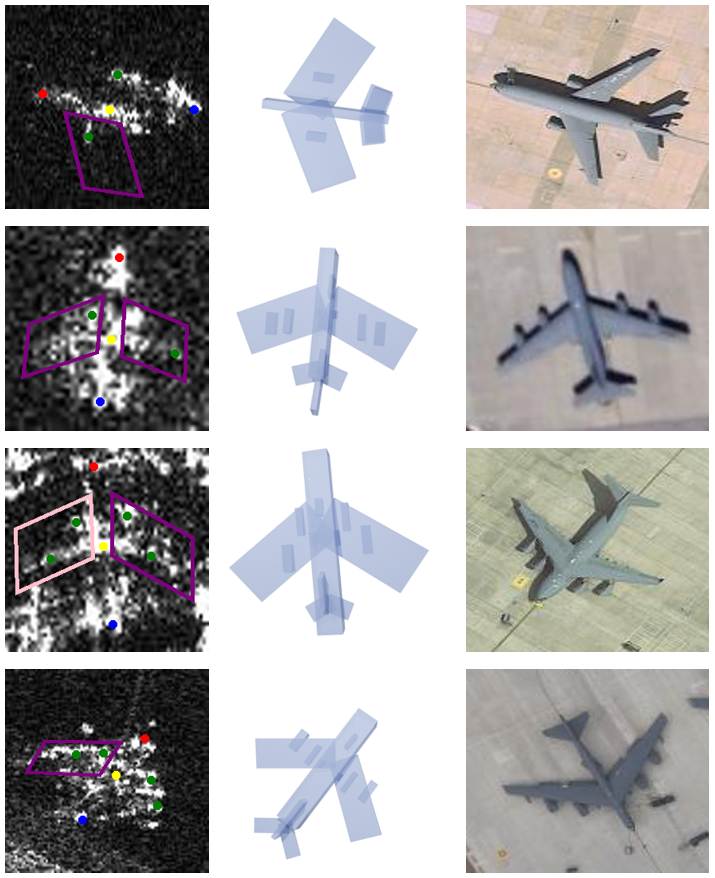}
	\caption{Structure recovery results of selected aircraft models and their optical remote sensing reference images.}\label{fig:real2}
\end{figure}

As mentioned in Section I, due to the lack of ground truth 3D models corresponding to real SAR data, a quantitative evaluation cannot be conducted in this section. Therefore, the reasonableness of reconstructed structure is qualitatively assessed based on the labeled aircraft models. As shown in Fig.~\ref{fig:real2}, the first and second columns show the keypoints and structure outputs from Step 1 and Step 2, respectively, while the third column shows optical remote sensing reference images of the corresponding aircraft. As observed, the reconstructed results keep consistent with the real shape in terms of structure (such as number of engines, component connections, etc.). The overall size of aircraft, as well as proportions and layout of components, also match reasonably well.

The results indicate that the structure recovery model trained on simulated data in Step 2 is robust and can effectively link with the keypoint detection model trained on real data in Step 1, thereby successfully performing structure recovery on real aircraft data.


\section{Conclusion and Discussion}

In this paper, we propose a novel SAR target structure recovery task. Unlike traditional methods that focus on target surface reconstruction or scattering center representations, we model SAR images of similar targets as a collection of observations that implicitly capture structural consistency and geometric diversity. The 3D structure of the target is described using OBBs and adjacency and symmetry relationships between them. To address the domain gap between simulated and real SAR images, we introduce a novel two-step algorithm based on the mapping-bridge of structural descriptors: the first step detects 2D keypoint information from aircraft SAR images using a multi-task learning framework and adaptive training strategy, and the second step uses a dual-stream GNN encoder and a RvNN decoder to infer hierarchical structure of aircraft from the keypoint information.

In the experiments, for Step 1’s real image keypoint detection, results show that the proposed multi-task learning algorithm outperforms individual task-based training. For Step 2’s structure recovery with simulated data, we introduce a subtree matching score (SMS) metric to evaluate the hierarchical structure reconstruction, and the evaluations show that the algorithm outperforms ablation baselines in terms of OBB geometric reconstruction accuracy and hierarchy consistency. For structure recovery with real SAR data, the test results demonstrate generalization ability of the proposed two-step algorithm, which can reliably recover highly reasonable 3D structures of aircraft in real-world scenarios.

This paper emphasizes analyzing SAR target images from structural and semantic perspectives, providing a new approach for SAR 3D reconstruction. However, several limitations remain and should be explored in future:
\subsubsection{Geometric completion and 3D reconstruction}  
The current work focuses on structure recovery, where components are represented by simple OBBs, which do not fully express complex geometric details of the targets. Future work can build upon this approach to fill in the internal geometry of each OBB, achieving more comprehensive 3D reconstruction.
    
\subsubsection{Integration with physical models}  
Structural-aware SAR image simulation and editing could be explored. By combining the physical process of SAR imaging with structured modeling, more realistic and controllable SAR images could be generated. Hierarchical representations could be used to adjust the local scattering characteristics of SAR images, simulating variations in material types or target layouts.
    
\subsubsection{Incorporation of viewpoint information}  
The current Step 2 algorithm does not consider viewpoint diversity when synthesizing simulated data, instead training on aircraft inputs with fixed orientations. Future work could incorporate multi-view variations as a data augmentation technique to further enhance the algorithm's generalization ability.
    
\subsubsection{Expanding the scope of study objects}  
Due to data limitations, the current study is focused on aircraft targets. Future research could explore applying the method to a wider range of target types (such as buildings, vehicles, etc.) to validate its generalizability and expand the application scenarios.

\small
\bibliographystyle{IEEEtranN}
\bibliography{main}

\end{document}